\documentclass[runningheads]{llncs}
\usepackage{graphicx}
\usepackage{comment}
\usepackage{amsmath,amssymb} 
\usepackage{color}

\usepackage[width=122mm,left=12mm,paperwidth=146mm,height=193mm,top=12mm,paperheight=217mm]{geometry}

\usepackage{microtype}
\usepackage{booktabs} 
\usepackage{enumitem}
\usepackage{float}
\usepackage{subfloat}
\usepackage{subfig}
\usepackage{multirow}
\usepackage{array}
\usepackage{makecell}
\usepackage{tabularx}
\usepackage[english]{babel}

\usepackage{tabularx}
\newcolumntype{R}{>{\raggedleft\arraybackslash}X}
\newcolumntype{L}{>{\raggedright\arraybackslash}X}
\newcolumntype{C}{>{\centering\arraybackslash}X}

\usepackage[ruled, lined, linesnumbered]{algorithm2e}

\graphicspath{{figures/}}

\begin{document}

\setlength{\abovedisplayskip}{4pt}
\setlength{\belowdisplayskip}{4pt}
\setlength{\abovedisplayshortskip}{0pt}
\setlength{\belowdisplayshortskip}{0pt}

\title{
Image Generation Via Minimizing Fr\'{e}chet Distance in Discriminator Feature Space\thanks{This work was done when several of the authors were in Criteo Inc., Palo Alto, USA.}
}



\titlerunning{Fr\'{e}chet-GAN}
%
\author{Khoa D. Doan\inst{1}
Saurav Manchanda\inst{2}
Fengjiao Wang \inst{3}
\and
Sathiya Keerthi \inst{4}
\and
Avradeep Bhowmik \inst{3}
\and
Chandan K. Reddy \inst{1}
}
\authorrunning{Doan K. et al.}
%

\institute{Virginia Tech, Arlington, USA \\ \email{khoadoan@vt.edu,reddy@cs.vt.edu}
\and University of Minnesota, USA \\ \email{manch043@umn.edu}
\and Amazon, USA 
\\ \email{w.fengjiao@gmail.com,avradeep.1@gmail.com}
\and Linkedin Corporation, USA \\ \email{keselvaraj@linkedin.com}
}
\maketitle

\begin{abstract}

For a given image generation problem, the intrinsic image manifold is often low dimensional. We use the intuition that it is much better to train the GAN generator by minimizing the distributional distance between real and generated images in a small dimensional feature space representing such a manifold than on the original pixel-space. We use the feature space of the GAN discriminator for such a representation. For distributional distance, we employ one of two choices: the Fr\'{e}chet distance or direct optimal transport (OT); these respectively lead us to two new GAN methods: Fr\'{e}chet-GAN and OT-GAN. The idea of employing Fr\'{e}chet distance comes from the success of Fr\'{e}chet Inception Distance as a solid evaluation metric in image generation. Fr\'{e}chet-GAN is attractive in several ways. We propose an efficient, numerically stable approach to calculate the Fr\'{e}chet distance and its gradient. The Fr\'{e}chet distance estimation requires a significantly less computation time than OT; this allows Fr\'{e}chet-GAN to use much larger mini-batch size in training than OT. More importantly, we conduct experiments on a number of benchmark datasets and show that Fr\'{e}chet-GAN (in particular) and OT-GAN have significantly better image generation capabilities than the existing representative primal and dual GAN approaches based on the Wasserstein distance.


\end{abstract}



\section{Introduction}

There has been a great amount of interest in generative modeling in recent years. This is driven largely by recently proposed generative models such as Generative Adversarial Networks (GANs) and Variational Autoencoders (VAEs). Different from classical generative models which directly sample from the high-dimensional space~\cite{blei2003latent,salakhutdinov2007restricted}, GANs and VAEs sample from a low-dimensional space and rely on deep neural networks to transform these samples into the high-dimensional target space. While VAEs have an appealing probabilistic interpretation, in the image domain their generated samples are often overly smooth. GANs are known to generate more realistic visual images, but training them has several challenges, most notably mode collapse and vanishing gradient.

The original GAN, proposed by \cite{goodfellow2014generative}, suffers from both frequent mode collapse and vanishing gradient. To address such problems, \cite{arjovsky2017towards,martin2017wasserstein} proposes to replace the Jensen-Shannon divergence objective in the original GAN with the Wasserstein distance objective. Wasserstein distance has a weaker topology assumption; thus the generated distribution more easily converges to the true distribution than in the case of some other divergences, without the mode collapse and gradient saturation problems in training. However, estimating the Wasserstein distance is a nontrivial task. Generally, there exist two main classes of approaches: 1) from the dual domain and 2) from the primal domain. 

Because of the intractability of estimating the Wasserstein distance from the primal domain, \cite{martin2017wasserstein} proposes to estimate the Wasserstein distance by employing Kantorovich-Rubeinstein duality and parameterizes the space of all 1-Lipschitz functions by a fixed network (critic), whose parameters are clamped to ensure the required Lipschitz smoothness.  Later, WGAN-GP~\cite{gulrajani2017improved} and WGAN-SN~\cite{miyato2018spectral} constrains the Lipschitz smoothness by using gradient-clipping regularization and gradient normalization, respectively. However, these approaches can have a significant approximation error when the learnable function space, which is controlled by their hyperparameters, is too large or too small. Furthermore, the dual domain requires the minimax setup, which is generally hard to optimize with problems such saddle-point. 

Alternatively, from the primal domain, \cite{iohara2018generative} estimates the empirical Wasserstein distance directly from the samples by solving the Optimal Transport (OT) problem. While solving the OT problem is a valid and promising direction, the computational cost, which is $O(N^{2.5}\log{N})$ where $N$ is the number of samples, is expensive. The empirical Wasserstein distance is also known to have high-variance and exponential sample complexity~\cite{deshpande2018generative,deshpande2019max,genevay2017learning}. Furthermore, minimizing the OT cost on the pixel space does not work for high-dimensional images~\cite{iohara2018generative}. 

To address the exponential sample-complexity problem, Sliced Wasserstein GAN (SWG)~\cite{deshpande2018generative} proposes to estimate the Sliced-Wassserstein distance, which randomly projects the data into many one-dimensional directions and approximates the Sliced-Wasserstein distance by averaging the one-dimensional Wasserstein distances along these directions. SWG has a polynomial sample-complexity~\cite{deshpande2018generative}. However, Sliced-Wasserstein distance is a very different metric, thus it does not have all the nice geometrical properties of Wasserstein distance~\cite{genevay2017learning}. Furthermore, in the high-dimensional space, a random one-dimensional direction will not likely lie on the data manifold, thus its Wasserstein distance is likely to be close to zero. For such a reason, SWG requires a large-number of random projections, which defeats its computational advantage as compared to OT. 
Later, to address this problem of SWG, Max-Sliced Wasserstein GAN (Max-SWG) proposes to find the best direction and minimize the distance along this direction~\cite{deshpande2019max}. Max-SWG shows a significant improvement over SWG while having a lower computational cost and memory footprint, and it is an important baseline to compare against. 

Our work started as an effort to improve the primal Wasserstein distance approach. Working on the raw, high dimensional image pixel space is clearly inappropriate. Most image generation problems have an intrinsic image manifold that is low dimensional and so it is better to work with a suitable low dimensional feature space that is problem/dataset specific. The main novelty of this paper comes from the use of the feature space of the GAN discriminator (that is designed to distinguish real and generated images) for this purpose. 

Having settled the feature space to work with, we also explore the type of distributional distance to employ. We could simply use the Wasserstein distance computed by OT. This leads us to our OT-GAN method. Alternatively, motivated by the success of Fr\'{e}chet Inception Distance as a solid evaluation metric in image generation problems, we also try the Fr\'{e}chet distance in the discriminator feature space. We call the resulting method as Fr\'{e}chet-GAN. We develop all the algorithmic and design details associated with Fr\'{e}chet-GAN and demonstrate that it is attractive in several ways, including a strong performance on several image generation benchmarks. 

The main contributions of our paper can be summarized as follows.
\begin{itemize}
    \item We propose Fr\'{e}chet-GAN and OT-GAN as novel and highly performing alternatives for existing GAN methods.
    \item We develop all the algorithmic and design details needed for their efficient implementation, e.g., the computation of the distances and their gradients. 
    \item We demonstrate the superiority of the proposed models, especially of Fr\'{e}chet-GAN, over other representative GAN approaches. Specifically, Fr\'{e}chet-GAN achieves a significant improvement in terms of both, the quality of the generated images and the Fr\'{e}chet Inception Distance metric, on the popular MNIST, CIFAR-10, CELEB-A and LSUN-Bedroom benchmark datasets.
\end{itemize}

The paper is organized as follows. In Section~\ref{sec:background}, we provide a review of the related primal GANs to our proposed models, noting the importance of minimizing the distributional distance in the low-dimensional feature space. In Section~\ref{sec:approach}, we describe the details of our proposed methods. Then, we present the quantitative and qualitative experimental results in Section \ref{sec:experiments}. Finally, Section~\ref{sec:conclusion} gives the conclusion.

\section{Background} \label{sec:background}

We begin by summarizing two related approaches of estimating the Wasserstein distance from the primal domain. Then, we discuss the importance of moving away from the pixel-space to a more suitable, low-dimensional feature space where the distributional distance is optimized. Finally, we discuss the Fr\'{e}chet Inception Distance, which is the motivation behind our Fr\'{e}chet-GAN model.

\subsection{Primal Wasserstein Distances} \label{subsec:related_ot}

In generative modeling, we learn to estimate the density $\mathbb{P}_d$ only from its empirical samples $\mathcal{D} = \{x|x \sim \mathbb{P}_d\}$. Different from the explicit generative models~\cite{salakhutdinov2007restricted,blei2003latent}, GANs model $\mathbb{P}_d$ as $\mathbb{P}_g$ implicitly through a mapping function from a low-dimensional latent space $\mathbb{P}_z$. Most GANs can be viewed as minimizing the distance between $\mathbb{P}_d$ and $\mathbb{P}_g$. One interesting distance of measures is the Wasserstein distance. Wasserstein distance has a weaker topology assumption and is the de facto metric for comparing distributions with non-overlapping supports. Minimizing the Wasserstein distance allows the generated distribution to more easily converge to the true distribution than in the case of other divergences, such as Jensen-Shannon divergence. Given the $L^p$ distance function $d(x, y)$, the Wasserstein-p distance between two distributions  $\mathbb{P}_d$ and $\mathbb{P}_g$ is defined as follows:
\begin{align}
        W_p(\mathbb{P}_d, \mathbb{P}_g) = \inf_{\gamma \in \Pi(\mathbb{P}_d, \mathbb{P}_g)} \int_{(x,y) \sim \gamma} p(x,y) d^p(x,y) dx dy
\end{align}
where $\Pi(\mathbb{P}_d, \mathbb{P}_g)$ is the set of all joint distributions of $x$ and $y$ whose marginals are $\mathbb{P}_d$ and $\mathbb{P}_g$, respectively.

Because directly estimating the Wasserstein distance is highly intractable, \cite{martin2017wasserstein} employs Kantorovich-Rubinstein duality to estimate the Wasserstein-1 distance as follows:
\begin{align}
        W_1(\mathbb{P}_d, \mathbb{P}_g) = \sup_{||f||_L \le 1} E_{x \sim \mathbb{P}_g}[f(x)] -  E_{x \sim \mathbb{P}_d}[f(x)]
\end{align}
where $f$ is a 1-Lipschitz continuous function. In practice, the function $f$ can be represented by a neural network. Enforcing the Lipschitz smoothness is still an open problem. Popular strategies such as weight clamping, gradient clipping and gradient normalization are employed~\cite{martin2017wasserstein,gulrajani2017improved,miyato2018spectral}. However, these techniques suffer from the Lipschitz-smoothness approximation in the parameterized function space. Consequently, training these models remain inefficient in practice because of their additional hyperparameter tuning.

Recently, \cite{iohara2018generative} proposes to directly estimate the Wasserstein-p distance from its primal domain by solving the OT problem from the empirical samples. Let $\mathcal{D}$ and $\mathcal{F}$ be the empirical samples which are drawn from $\mathbb{P}_d$ and $\mathbb{P}_g$ respective. Given a sample $x_i$ from $\mathcal{D}$ and a sample $y_j$ from $\mathcal{F}$, the $L^p$ distance function is parameterized by the generator's parameters $\Theta_G$ and denoted as $d_{\Theta_G}(x_i, y_j)$. The OT cost between $\mathcal{D}$ and $\mathcal{F}$, is calculated by solving the following linear-sum assignment problem:
\begin{equation}
C(\mathcal{D}, \mathcal{F}) = \min_{\Theta_G} \sum_i^I \sum_j^J m_{i,j} d_{\Theta_G}^p(x_i, y_j)
\label{eqn:primal_empirical}
\end{equation}
subject to the constraints:
\begin{align}
\sum_i m_{i,j} = 1 \; \forall j, \;\;\; \sum_j m_{i,j} = 1 \; \forall i, \;\;\; m_{i,j} \in \{0, 1\} \; \forall i, j
\end{align}
Note that the matrix $M$ is a doubly stochastic matrix and the solution is a permutation matrix.
Because this method does not have the generator-discriminator setup, it is no longer a GAN. Clearly, minimizing OT on the high dimensional image pixel space is inappropriate. As shown in Figure~\ref{tbl:generated_images_pixel_space}, the models generate CIFAR-10 images (higher-dimensional) with significant mode collapse or white noise than the generated MNIST images (low-dimensional). Minimizing OT on the original input space only works in domains where the input data is already low-dimensional and the Frobenius distance is suitable~\cite{manchanda2020regression,doan2020hashing}. Another challenge of using OT is its high computational cost. Solving the linear-sum assignment program has a complexity of $O(N^{2.5}\log{N})$ where $N$ is the number of samples~\cite{burkard2009assignment}. Reducing OT's computational cost is necessary to utilize it in practice~\cite{doan2020hashing}.

\renewcommand\tablename{Figure}
\begin{table*}[t]
\centering
\begin{tabular}{b{1em}cccc}
& \multicolumn{2}{c}{\textbf{SWG}} & \multicolumn{2}{c}{\textbf{OT-GAN}} \\

& feature space & pixel space &  feature space & pixel space \\

\rotatebox{90}{\textbf{MNIST}} & \includegraphics[width=1.1in]{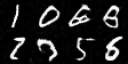} 
&
\includegraphics[width=1.1in]{figures/pixel-space/mnist-SWG-withD-2r4c.png}
&
\includegraphics[width=1.1in]{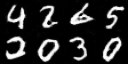} 
&
\includegraphics[width=1.1in]{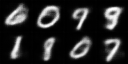} \\

\rotatebox{90}{\textbf{CIFAR-10}} & 
\includegraphics[width=1.1in]{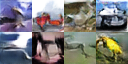} 
&
\includegraphics[width=1.1in]{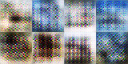}
&
\includegraphics[width=1.1in]{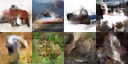} 
&
\includegraphics[width=1.1in]{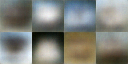} \\
 
\end{tabular}
\caption{Generated images when minimizing the divergences (Sliced Wassertein distance and OT) on the pixel versus feature space.}
\label{tbl:generated_images_pixel_space}
\vspace{-20pt}
\end{table*}


Along the same primal direction, SWG estimates the Sliced Wasserstein distance, which is also a valid distance metric between two distributions~\cite{deshpande2018generative}. Like OT, in principle, SWG too does not need the generator-discriminator setup. However, as shown in Figure~\ref{tbl:generated_images_pixel_space}, SWG fails to generate meaningful images when it minimizes the Sliced Wasserstein distance in the pixel space. For high-dimensional datasets, SWG transforms the high-dimensional input space into a suitable low-dimensional space. The low-dimensional space is chosen to maximally separates the real and fake samples. Thus, SWG has the generator-discriminator setup, where the discriminator is trained to classify real and fake samples while the generator is trained to minimize the Sliced Wasserstein distance on the projected low-dimensional space. Suppose $D$ is the discriminator and some intermediate output of $D$ is defined as $D'$. The $D'(x)$ space is called the \textbf{feature space}.  SWG learns to generate images by having the following objectives:
\begin{align}
    & \min_{\Theta_D} E_{x \sim \mathbb{P}_d}[-log(D(x)] + E_{x \sim \mathbb{P}_g}[-log(1-D(x))] \\
    & \min_{\Theta_G} \frac{1}{|\Omega|} \sum_{\omega \in \Omega} W_2^2 (D'(\mathcal{D})\omega, D'(\mathcal{F})\omega)
\end{align}
where $W^2_2$ is the one-dimensional Wasserstein-2 distance in the direction defined by $\omega$ and $\Omega$ is the set of one-dimensional random projections. The one-dimensional Wasserstein distance has a trivial computation~\cite{deshpande2018generative}. While SWG works well in many complex generative modeling tasks, the random projections loose a lot of information even in the feature space $D'(x)$. This is because a randomly projected direction is unlikely along the data manifold. Consequently, SWG needs a large number of random projections to work well, which defeats its computational advantage over OT. Specifically, SWG estimates the Wasserstein distance with a computational cost of $O(KN\log(N))$, whose computation complexity is better than the complexity of OT only when $K$ is smaller than $N^{1.5}$. However, in practice, the number of random directions $K$ is often larger than $N^{1.5}$, . For example, in~\cite{deshpande2018generative}, for a mini-batch size of $64$, SWG needs $K=10,000$ projections, which is significantly larger than $64^{1.5}$, to generate high-dimensional images. To address this problem, Max-SWG finds the best direction and estimates the Wasserstein distance along this direction~\cite{deshpande2019max}. Max-SWG significantly reduces the computational complexity and memory footprint of SWG. Furthermore,  Max-SWG can generate images with high visual quality.

\subsection{Fr\'{e}chet distance} \label{subsec:related_frechet_distance}

Given two multivariate Gaussian distributions $\mathbb{P}_g$ and $\mathbb{P}_d$ with means $\mu_g$ and $\mu_d$ respectively, and covariances $\Sigma_g$ and $\Sigma_d$ respectively, the Fr\'{e}chet distance between  $\mathbb{P}_g$ and $\mathbb{P}_d$ is defined as follows:
\begin{align}
    \text{FD}(\mathbb{P}_g, \mathbb{P}_d) = ||\mu_d - \mu_g||^2 + Tr \left(\Sigma_d + \Sigma_g - 2\sqrt{\Sigma_d\Sigma_g}\right)
\end{align}
In the image domain, Fr\'{e}chet distance is popularly used to calculate the Fr\'{e}chet Inception Distance (FID) of the generative models. FID is calculated by first extracting the activations of the real and generated images on pool3 layer of the pre-trained InceptionV3 network, then computing the Fr\'{e}chet distance from the extracted features of the samples~\cite{heusel2017gans}. The InceptionV3 network is pre-trained on the ImageNet dataset. Lower FID is generally better and the corresponding generative model is able to generate images with highly realistic visual quality. Interestingly, the Fr\'{e}chet distance is equivalent to the Wassertein-2 distance of the two multivariate Gaussian distributions. 

\section{Approach} \label{sec:approach}

We consider the problem of improving the existing primal Wasserstein distance approaches in the image generation domain. First, we discuss how to efficiently scale up OT to the high-dimensional image cases using a more suitable, low-dimensional feature space. Then, we propose a novel, computationally efficient alternative to OT. This involves calculating the Fr\'{e}chet distance between two distributions in the feature space.

\subsection{Scaling up OT} \label{subsec:scaling_ot}

While minimizing OT in the pixel space is able to generate MNIST digits, the digits are overly smooth (Figure~\ref{tbl:generated_images_pixel_space}), similar to those generated by VAE. We conjecture that this is because of the lack of the discriminator, which discriminates samples as real or fake. Specifically, minimizing the Wasserstein distance in the pixel-space causes the algorithm to easily fall into a local minimum where the generated distribution $\mathbb{P}_g$ is the average of other possible distributions that also represent realistic digits. Thus, the discriminator is necessary to penalize the distribution whose digits are visually smooth but unrealistic. Beyond MNIST, OT also fails to generate higher-dimensional images such as CIFAR-10 and CELEB-A. It is known that the Wasserstein distance has an exponential sample complexity~\cite{deshpande2019max}: higher-dimensional data need an impractical amount of samples to accurately estimate the Wasserstein distance. While natural images have high dimensions, the intrinsic manifold often lies on a much lower-dimensional subspace. Therefore, directly estimating OT on the pixel-space is highly ineffective.  For these reasons, in this paper, we argue that the discriminator is necessary to have OT work effectively on a variety of image datasets. 

Inspired by SWG, we employ a discriminator whose primary objective is to find a low-dimensional subspace that best discriminates the real and fake samples. Specifically, suppose that $D$ is the discriminator function and $D'$ is the part of $D$ which we consider as the \textbf{feature space}. Instead of minimizing OT between $\mathbb{P}_g$ and $\mathbb{P}_d$, we approximate OT on the output distributions of $D'$; that is, we minimize OT on the distribution of $D'(\mathcal{D})$, denoted as $\mathbb{P}_d^{D'}$, and the distribution of $D'(\mathcal{F})$, denoted as $\mathbb{P}_g^{D'}$. In other words, given the generator's parameters $\Theta_G$ and the discriminator's parameters $\Theta_D$, we have the following objectives that are independently optimized:
\begin{align}
    & \min_{\Theta_D} E_{x \sim \mathbb{P}_d}[-log(D(x)] + E_{x \sim \mathbb{P}_g}[-log(1-D(x))] \label{eqn:d_loss_real_fake} \\
    & \min_{\Theta_G} C(D'(\mathcal{D}),D'(\mathcal{F})) \label{eqn:g_loss_OT}
\end{align}
We call this model OT-GAN. Figure~\ref{tbl:generated_images_pixel_space} shows the effectiveness of OT-GAN. The generated images, in both MNIST and CIFAR-10 examples, have a significant improvement over the pixel-space minimization.  Note that, OT-GAN does not have the minimax setup because the objective functions of the generator and discriminator are different. 

\subsection{GAN based on Fr\'{e}chet distance}


While OT-GAN offers a significant improvement over its pixel-space minimization, the high computational complexity is prohibitive in practice. In this section, we propose a computationally efficient alternative that is based on the Fr\'{e}chet distance.

Assuming the feature space $\mathbb{P}_d^{D'}$ and $\mathbb{P}_g^{D'}$ are multivariate Gaussian distributions, we can approximate the Wasserstein-2 distance via the Fr\'{e}chet distance, as described in Section~\ref{subsec:related_frechet_distance}. The means of $\mathbb{P}_d^{D'}$ and $\mathbb{P}_g^{D'}$ are $\mu_d^{D'}$ and $\mu_g^{D'}$, respectively, while their covariances are $\Sigma_d^{D'}$ and $\Sigma_g^{D'}$, respectively. Note that only $\mu_g^{D'}$ and $\Sigma_g^{D'}$  are functions of the generator's parameters $\Theta_G$. The Fr\'{e}chet distance of the multivariate Gaussians $\mathbb{P}_d^{D'}$ and $\mathbb{P}_g^{D'}$ is defined as:
\begin{align}
    \text{FD}( \mathbb{P}_d^{D'}, \mathbb{P}_g^{D'}) = ||\mu_d^{D'} - \mu_g^{D'}||^2 + Tr\left(\Sigma_d^{D'} + \Sigma_g^{D'} - 2\sqrt{\Sigma_d^{D'}\Sigma_g^{D'}}\right)
    \label{eqn:frechet_distance_feature_space}
\end{align}

We now describe an efficient algorithm for computing the Fr\'{e}chet distance. While the Fr\'{e}chet distance can be analytically calculated, we need to pay special attention to avoid numerical errors, especially when training the model on conventional computing hardware. In Equation~\ref{eqn:frechet_distance_feature_space}, computing the values and derivatives (w.r.t $\Theta_G$) of $||\mu_d^{D'}-\mu_g^{D'}||^2$ and $Tr(\Sigma_g^{D'})$ is trivial.  Therefore, we focus the discussion of this section to computing the matrix square root term $\sqrt{\Sigma_d^{D'}\Sigma_g^{D'}}$ and its derivative  $\partial_{\Theta_G} \sqrt{\Sigma_d^{D'}\Sigma_g^{D'}}$. Note that $\Sigma_d^{D'}\Sigma_g^{D'}$ is a $d \times d$ matrix, where $d$ is the dimension of the feature space $D'$.

In the following discussion, we rely on some useful mathematical results, which will be provided in the Appendix of this paper.

\subsubsection{Computing the matrix square root:} \label{subsubsec:computing_squareroot}

Suppose $A \in \mathcal{R}^{d \times d}$ is a positive semi-definite (PSD) matrix and its SVD is given by $S \Sigma S^T$, where $\Sigma$ is a diagonal matrix and $S$ is an orthogonal matrix. Note that the diagonal entries of $\Sigma$ are non-negative real numbers. Define the matrix $B$ as follows $B = S \sqrt{\Sigma} S^T$. $B$ is a PSD matrix and $\sqrt{\Sigma}$ has non-negative diagonal entries. It is trivial to see that $B$ is the square root matrix of $A$. Furthermore, it can be shown that $B$ is the unique square root of the symmetric, PSD matrix $A$.




\subsubsection{Computing the matrix derivative:} 

Given the PSD matrix $A \in \mathcal{R}^{d \times d}$ and its square root matrix $B \in \mathcal{R}^{d \times d}$, we can derive the following equation by applying implicit differentiation on the equation $A = BB$:
\begin{align}
\partial_{A} = \partial{B} \times B + B \times \partial{B}
\end{align}
It turns out that this equation has the same form as the Sylvester equation, and there is a closed-form solution for $\partial{B}$, as follows:
\begin{align}
    \text{vec}(\partial{B}) = (B^T \otimes B)^{-1} \text{vec}(\partial{A})
    \label{eqn:sylvester_equation}
\end{align}
where $\text{vec}(X)$ is a vectorization operation that stacks columns of $X$ into a column vector, and $\otimes$ is the Kronecker product. However, this is a naive solution with a high computational cost of $O(d^3)$. In practice, we can also employ the Bartels-Stewart algorithm to solve this equation~\cite{bartels1972}. The Bartels-Stewart algorithm has a computational complexity of $O(d^{1.5})$, which is a significant improvement over the naive solution.


In principle, if we can compute $\sqrt{A}$, we can compute its derivative, $\partial{\sqrt{A}}$. However, in practice,  ~\cite{ionescu2015matrix} suggests that the SVD is ill-conditioned when some of the eigenvalues of $A$ are close to each other. In our experiments, we also observe that the eigenvalues become negative and the gradients become unstable when this condition happens. Therefore, we propose to compute $\sqrt{A}$ using the Newton-Schultz iterative algorithm~\cite{higham1997stable}, which we describe next. The Newton-Schultz algorithm is an extension to the popular Denman-Beaver algorithm~\cite{denman1976matrix}, but is more efficient because it only involves matrix multiplications instead of calculation of inverses. Let $A$ be a normalized matrix, and let $Y_0 = A$ and $Z_0 = I$, where $I$ is the identity matrix. In each iteration, we update $Y_{t+1}$ and $Z_{t+1}$ as:
\begin{align}
    U = \frac{1}{2} (3I - Z_t Y_t), Y_{t+1} = Y_t U, Z = U Z_t  
    \label{eqn:newton_schultz}
\end{align}

When $t \rightarrow \infty$, the matrices $Y$ and $Z$ converge quadratically to $A^{1/2}$ and $A^{-1/2}$, respectively. Note that, when computing $\sqrt{A}$ iteratively, we can rely on automatic differentiation, which is supported in most modern deep learning frameworks, to compute its derivative. However, the memory overhead grows linearly with the number of iterations if the computation graph is preserved. Thus, to maintain both numerical accuracy and low-memory overhead, we suggest to first compute $\sqrt{A}$ iteratively without keeping the computation graph in memory, then solve the Sylvester equation to directly compute its gradient with respect to the parameters. In our experiments, we observe that 10-15 iterations  are adequate to converge to a good numerical numerical error in a conventional GPU (Figure~\ref{subfig-1:sqrt_errors}). Thus, for simplicity, we rely on the automatic differentiation of the iterative solution to calculate the gradient of $\sqrt{A}$. In Figure~\ref{subfig-2:sqrt_gradients}, we also observe that the gradient of the parameters is stable when we train using this approach.

\renewcommand\figurename{Figure}
\begin{figure}[!t]
\setcounter{figure}{1}
\vspace{-5pt}
 \centering
 \renewcommand{\thesubfigure}{a}
 \subfloat[Approximation Error \label{subfig-1:sqrt_errors}]{%
   \includegraphics[width=0.45 \textwidth]{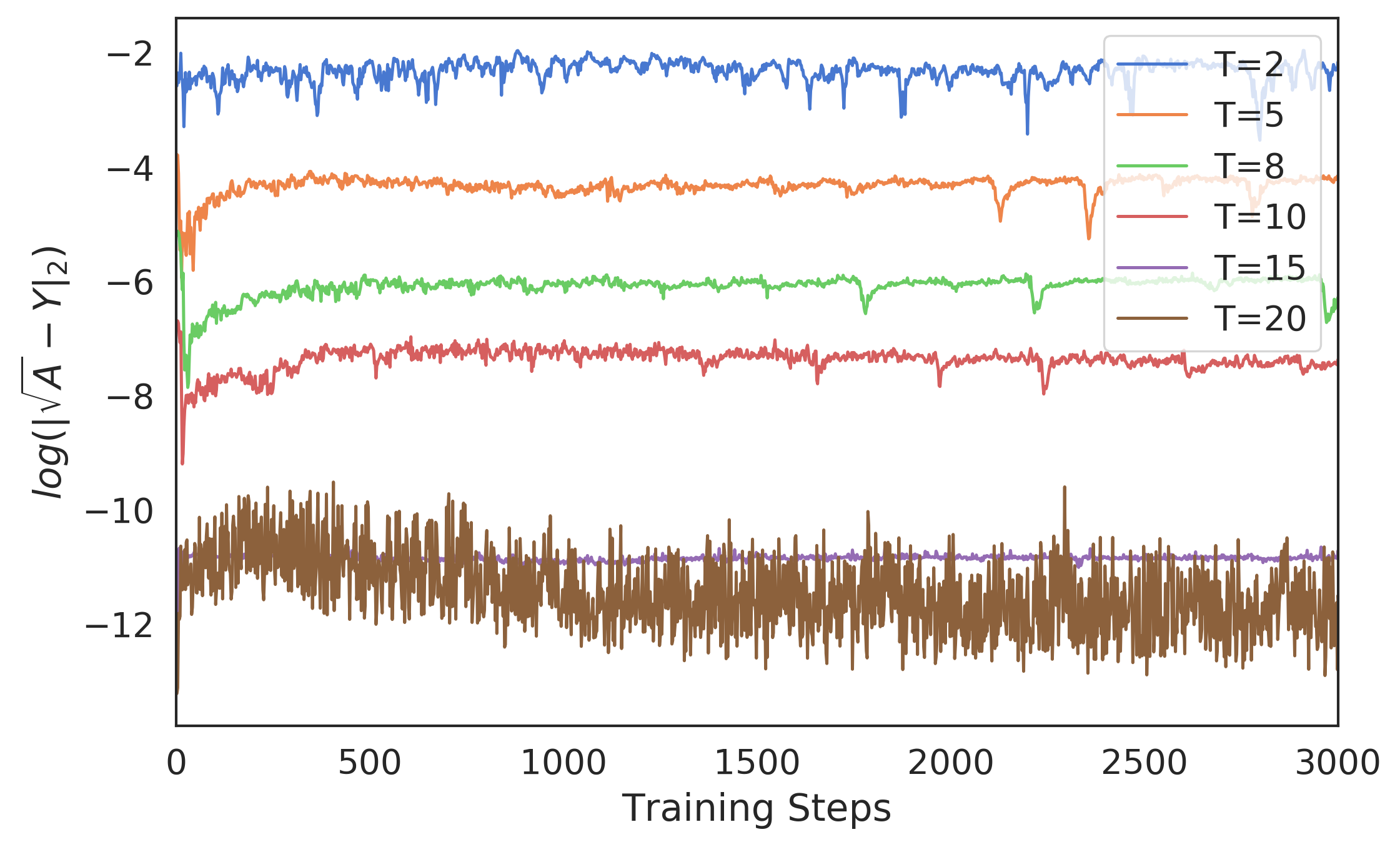}
 }
 \hfill
 \renewcommand{\thesubfigure}{b}
 \subfloat[Gradient Magnitude \label{subfig-2:sqrt_gradients}]{%
   \includegraphics[width=0.45 \textwidth]{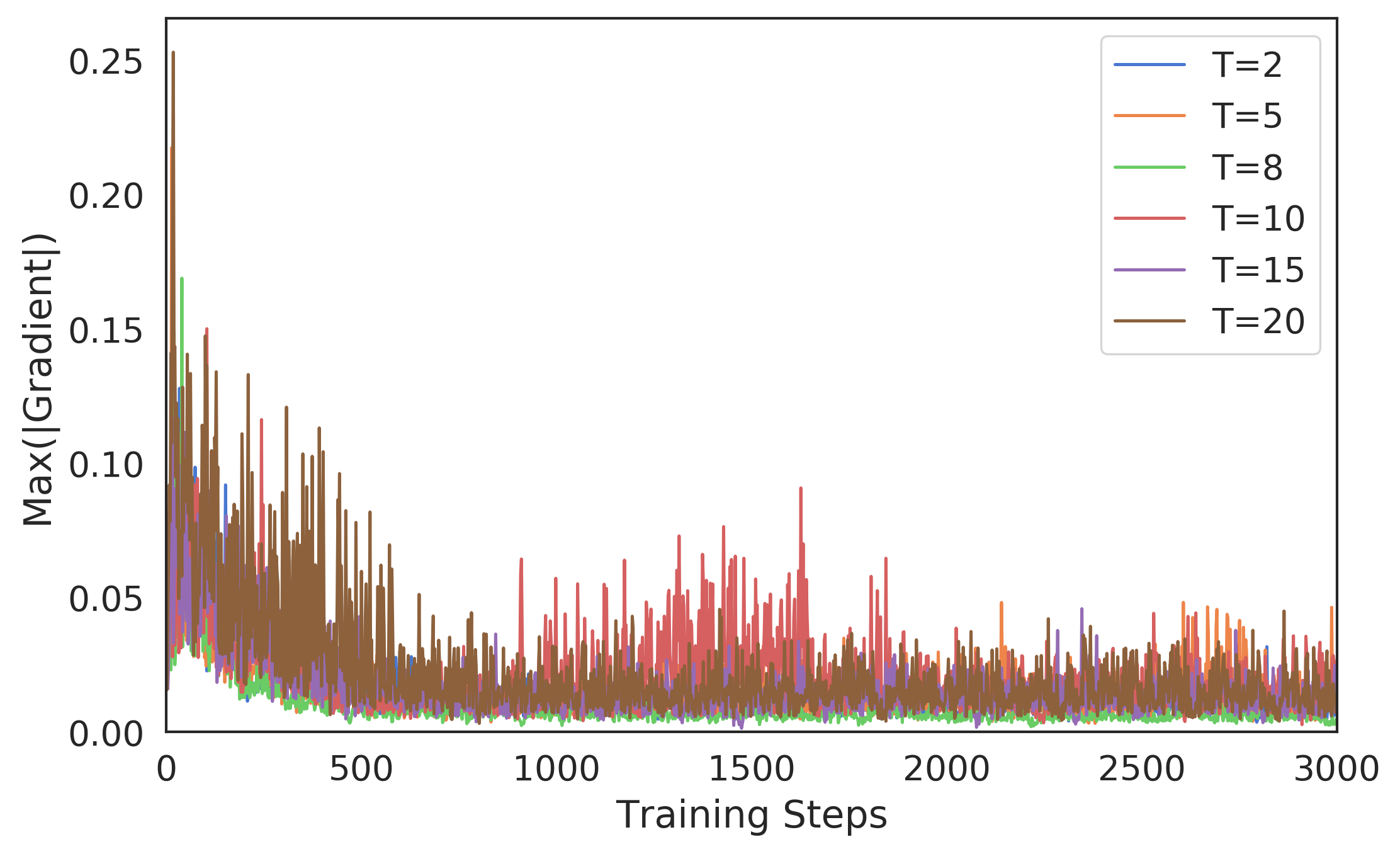}
 }
 \caption{Behavior of our proposed matrix square root calculation during training with different numbers of iterations $T$.}
 \label{fig:}
\vspace{-20pt}
\end{figure}

\subsubsection{Fr\'{e}chet-GAN}

While the pixel distributions $\mathbb{P}_g$ and especially $\mathbb{P}_d$ are not multivariate Gaussians, it is possible to enforce the feature space to be multivariate Gaussian, especially when the discriminator is a convolutional neural network. Inspired by the mechanics of calculating FID, we replace the last convolutional layer of $D'$ by an Average Pooling (AP) layer. Because of the effect of AP and the Central Limit Theorem, the output follows a multivariate Gaussian distribution. Heuristically, we find that using Max Pooling (MP) also yields multivariate Gaussian output in the modified DCGAN architecture~\cite{heusel2017gans}. Interestingly, the use of MP yields a significant better result than AP; Although, right now we do not have a precise explanation about this, we conjecture that MP results in a feature space where it is easier for the discriminator to discriminate real and generated examples. Note that the data covariance matrices $\Sigma_d^{D'}$ and  $\Sigma_g^{D'}$ are PSD.








We call the proposed approach Fr\'{e}chet-GAN, and the learning algorithm of Fr\'{e}chet-GAN is summarized in Algorithm~\ref{alg:final_algorithm}. The code of Fr\'{e}chet-GAN is available on Github\footnote{\url{https://github.com/khoadoan/Frechet-GAN}}. We alternately train the discriminator and the generator. In each iteration, we draw samples from the real and generated distributions $\mathbb{P}_d$ and $\mathbb{P}_g$, respectively. Then, we train the discriminator, for $k$ iterations, to learn the feature space that best discriminates the real and generated samples. To update the generator, we perform the forward pass to compute the Fr\'{e}chet distance in the feature space defined by $D'$. For the square root term $\sqrt{\Sigma_d^{D'}\Sigma_g^{D'}}$, we perform the forward pass using the proposed Newton-Schultz iterative algorithm and perform the backward pass by solving the Sylvester Equation, defined in Equation~\ref{eqn:sylvester_equation}, using the  Bartels-Stewart algorithm. In our implementation, we rely on automatic differentiation by keeping the computation graph of the iterative computation in memory.
        
        
        
        
 
\begin{algorithm}[!t]
\DontPrintSemicolon
\KwIn{The generator's parameters $\Theta_G$, the discriminator's parameters $\Theta_D$, the mini-batch size $N$, and the learning rate $\alpha$. }
\While{$\Theta_G$ is not converged}{
sample $N$ real samples $\mathcal{D} \sim \mathbb{P}_d$ and $N$ generated samples $\mathcal{F} \sim \mathbb{P}_g$;\;
    
      
    
    \Begin(compute the discriminator loss)
    {
      $L_D \leftarrow \sum_{x \in \mathcal{D}} [-log(D(x)] + \sum_{x \in \mathcal{F}}[-log(1-D(x))]$ \;
    }
    $\Theta_D \leftarrow \Theta_D - \alpha \nabla_{\Theta_D}{L_D}$.\;
    
    \Begin(compute the generator's loss)
    {
      compute the features  $D'(\mathcal{D})$ and $D'(\mathcal{F})$. \;
      compute $L_G$ in Equation~\ref{eqn:frechet_distance_feature_space};
      for $\sqrt{\Sigma_d^{D'} \Sigma_g^{D'}}$, compute its value using Newton Schultz iterations in Equation~\ref{eqn:newton_schultz} and its derivative by solving Equation~\ref{eqn:sylvester_equation}. \;
    }
    $\Theta_G \leftarrow \Theta_G - \alpha \nabla_{\Theta_G}{L}$.
}
\caption{Fr\'{e}chet-GAN Training.}\label{alg:final_algorithm}
\end{algorithm}
 
    

        
        

The computation cost of Fr\'{e}chet distance is largely dominated by the calculation of the covariance matrices and the square root term $\sqrt{\Sigma_d^{D'}\Sigma_g^{D'}}$. Specifically, the computation complexity using the non-iterative algorithm (using SVD) is $O(min(Nd^2, d^3))$, while using the Newton Schultz algorithm, it is $O(min(Nd^2, Td^2))$ where $T$ is the number of iterations. This is significantly smaller than OT's computational cost, especially when $N$ is large. This allows us to use a large mini-batch size to estimate the Fr\'{e}chet distance, which is prohibitive for OT.


\section{Experimental Results} \label{sec:experiments}

\begin{table*}[!t]




\renewcommand\tablename{Figure}
\centering
\begin{tabular}{c}
\includegraphics[width=4in]{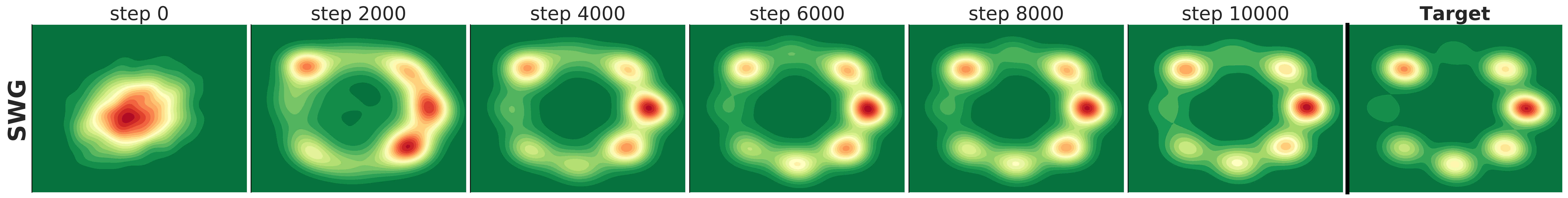} \\
\includegraphics[width=4in]{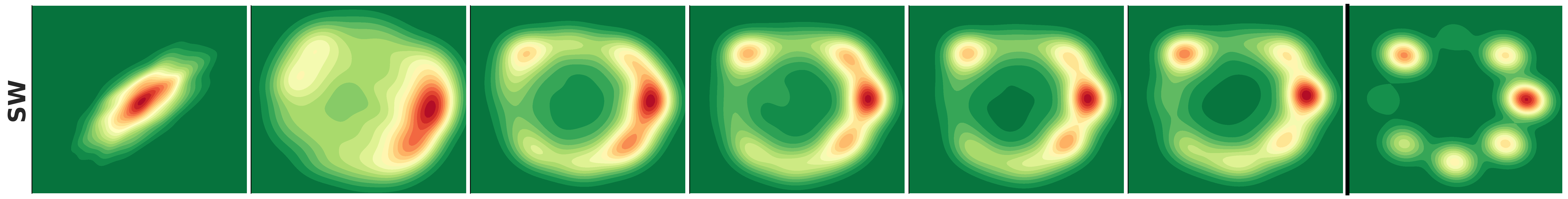} \\
\includegraphics[width=4in]{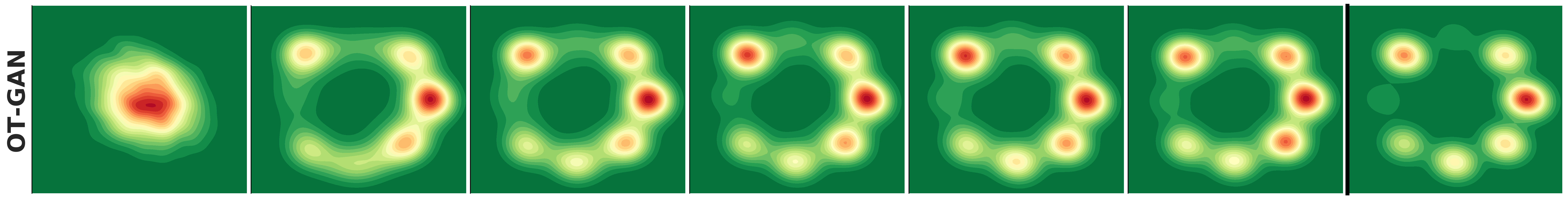} \\
\includegraphics[width=4in]{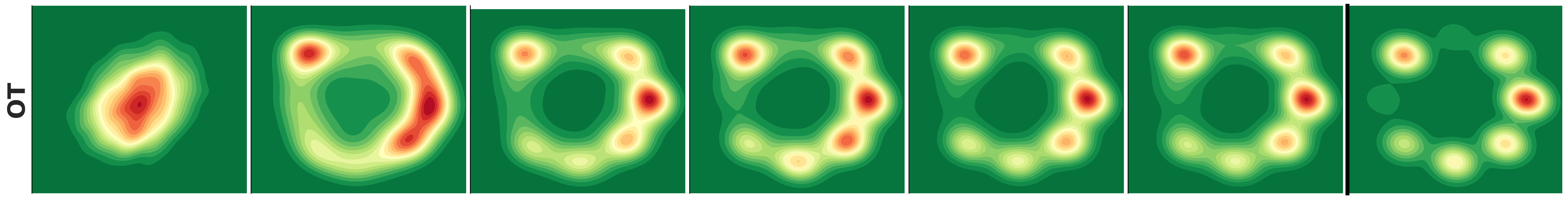} \\
\includegraphics[width=4in]{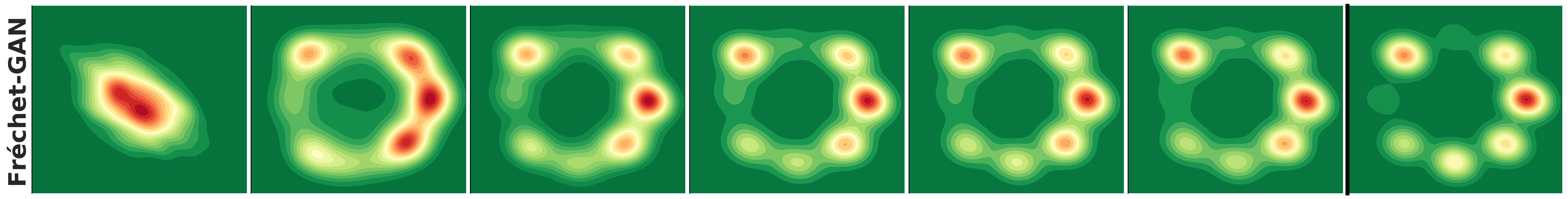} \\

\end{tabular}
\caption{Dynamic results of 8-Gaussian estimation. The final column is the real distribution. SW and OT refer to minimizing Sliced Wasserstein distance and OT, respectively, without the discriminator.}
\label{fig:synthetic_unrolled}

\vspace{-20pt}
\end{table*}

In this section, we present experimental results to demonstrate the effectiveness of our proposed methods, specifically OT-GAN and Fr\'{e}chet-GAN. We show both the qualitative and quantitative results on the image generation task on four different popular datasets: \textbf{MNIST} ~\cite{lecun1998mnist}, \textbf{CIFAR-10} ~\cite{krizhevsky2009learning}, \textbf{CELEB-A} ~\cite{liu2015deep}, and \textbf{LSUN-Bedroom}~\cite{yu2015lsun}. 

We compared OT-GAN and Fr\'{e}chet-GAN with the following models: WGAN~\cite{martin2017wasserstein}, WGAN-GP~\cite{gulrajani2017improved}, SWG~\cite{deshpande2018generative}, Max-SWG~\cite{deshpande2019max}. For all the methods, including ours, we use the DCGAN architecture. The use of the same network architecture allows us to objectively evaluate the capabilities of methods. Wherever possible, we use the same hyperparameter setting of the baselines as provided in the original papers. In other cases, we perform model selections based on FID, and report the results from the models' best configurations. FID is evaluated on 50,000 generated images and 50,000 real images. The best models are trained for 100 epochs on MNIST and CIFAR-10, 50 epochs on CELEB-A and 20 epochs on LSUN-Bedroom. 

\subsection{Performance with and without the discriminator}

In this section, we perform the experiment of generating the classic 8-Gaussian data~\cite{metz2016unrolled}, which has been previously used to study mode collapse. The generator and discriminator are two-hidden layer perceptions (MLPs) with Rectified-linear activation. Figure~\ref{fig:synthetic_unrolled} shows the generated samples during the training process. Without the discriminator, both SW and OT (directly minimizing Sliced Wasserstein distance and OT respectively on the original data space) exhibit signs of mode-collapse. However, when the discriminator is used, all models, including Fr\'{e}chet-GAN, can recover almost perfectly the original data. Furthermore, Fr\'{e}chet-GAN converges to the target distribution more quickly than the other GANs. This experiment demonstrates the effectiveness of minimizing the distributional distances in the feature space. 

\begin{table*}[!t]

\renewcommand\tablename{Figure}
\centering
\begin{tabular}{b{1em}cccc}
& \textbf{SWG} & \textbf{Max-SWG} & \textbf{OT-GAN} & \textbf{Fr\'{e}chet-GAN}  \\
    \rotatebox{90}{\textbf{Conv}} & \includegraphics[width=1in]{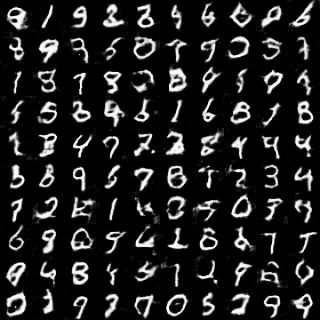} &
    \includegraphics[width=1in]{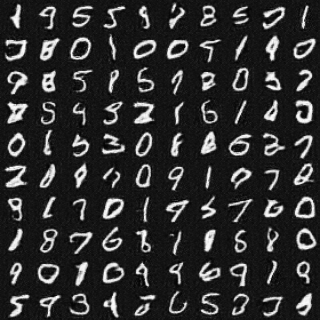} &
    \includegraphics[width=1in]{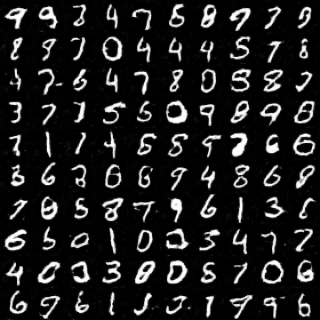} & \includegraphics[width=1in]{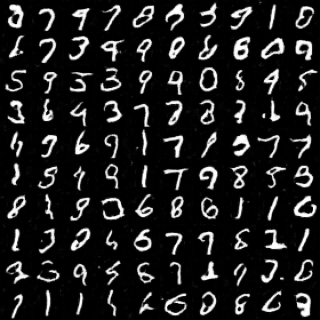} \\

    \rotatebox{90}{\textbf{Conv + BN}} & \includegraphics[width=1in]{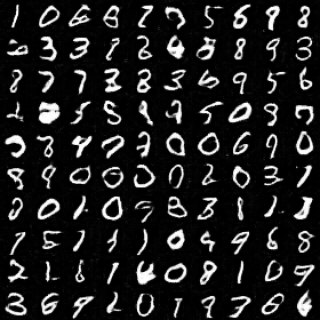} & 
    \includegraphics[width=1in]{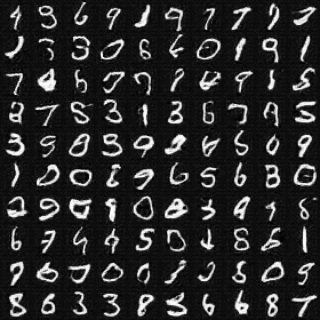} &
    \includegraphics[width=1in]{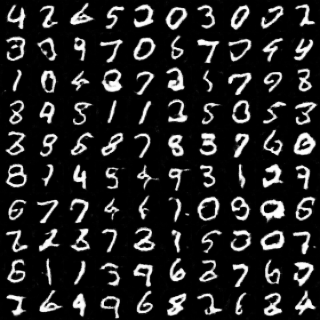} & \includegraphics[width=1in]{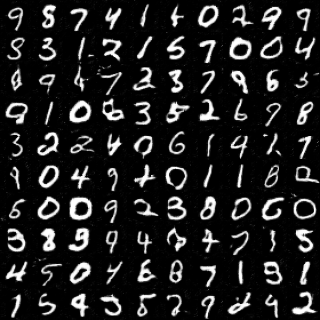} \\

 
 
\end{tabular}
\caption{Generated MNIST digits with and without BN in the DCGAN Architecture.}
\label{fig:generated_images_without_BN}
\vspace{-20pt}
\end{table*}

\renewcommand\tablename{Table}
\begin{table*}[!t]
\centering
\caption{FID with and without BN.}

\begin{tabularx}{\linewidth}{llCCCCCC}
\hline
\hline
& & WGAN & WGAN-GP & SWG & Max-SWG & OT-GAN & Fr\'{e}chet-GAN \\

\hline
\hline

\multirow{2}{*}{MNIST} & Conv              & 19.12 & 21.24 & 40.47 & 37.68 & 18.87 & 15.51  \\
                       & Conv + BN         & 18.86 & 20.35 & 17.16 & 38.63 & 13.13 & 21.22 \\ 
\hline

\multirow{2}{*}{CIFAR-10} & Conv            & 290.92 & 85.84  & 76.44 & 69.34 & 163.94 & 38.62  \\
                          & Conv + BN       & 44.51 & 36.24 & 26.68 & 23.56 & 32.50 & 24.64 \\
\hline

\end{tabularx}
\label{tbl:fids_nobn}
\vspace{-15pt}
\end{table*}

Note that, for Fr\'{e}chet-GAN, we simply calculate the Fr\'{e}chet distance directly on the last layer of the MLP discriminator. Although this layer's activation does not follow the multivariate Gaussian, Fr\'{e}chet-GAN still works. Reasons behind such a robust behavior of Fr\'{e}chet-GAN needs further investigation.




\subsection{Stability in generating images}

\cite{martin2017wasserstein} suggests to evaluate the training stability of GANs, especially the phenomenon of mode collapse, by removing the Batch Normalization layers (BN). In this Section, we present the results for generating MNIST and CIFAR-10 images when we use the Convolution layers with and without BN.

Figure~\ref{fig:generated_images_without_BN} shows the generated MNIST digits of two popular primal baselines, SWG and Max-SWG, as well as proposed GANs. Previously, primal GANs show significant improvement over  non-Wasserstein GANs and dual-Wasserstein GANs with respect to mode collapse~\cite{deshpande2018generative}. We observe that both OT-GAN and Fr\'{e}chet-GAN can generate meaningful digits regardless of the utilization of BN. 

\begin{table*}[!t]
\renewcommand\tablename{Figure}
\centering
\begin{tabular}{ccc}
\hline
\textbf{CIFAR-10 32x32} & \textbf{CELEB-A 64x64} & \textbf{LSUN-Bedroom 64x64} \\ \hline

\multicolumn{3}{c}{(a) Wasserstein GAN} \\
\includegraphics[width=1.3in]{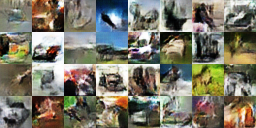} & \includegraphics[width=1.3in]{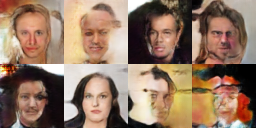} & \includegraphics[width=1.3in]{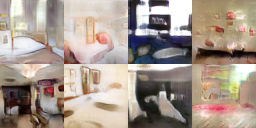} \\ \hline

\multicolumn{3}{c}{(b) Wasserstein GAN with Gradient Penalty} \\
\includegraphics[width=1.3in]{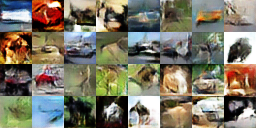} & \includegraphics[width=1.3in]{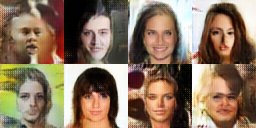} & \includegraphics[width=1.3in]{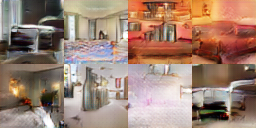} \\ \hline

\multicolumn{3}{c}{(c) Sliced Wasserstein GAN} \\

\includegraphics[width=1.3in]{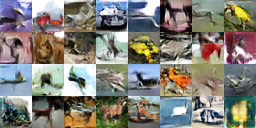} & \includegraphics[width=1.3in]{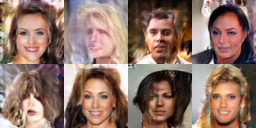} & \includegraphics[width=1.3in]{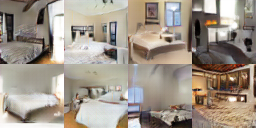} \\ \hline

\multicolumn{3}{c}{(d) Max-Sliced Wasserstein GAN} \\

\includegraphics[width=1.3in]{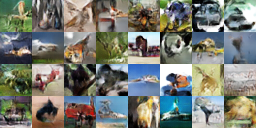} & \includegraphics[width=1.3in]{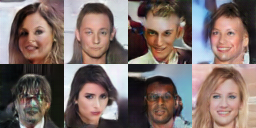} & \includegraphics[width=1.3in]{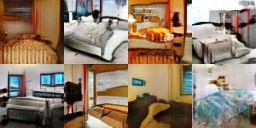} \\ \hline

\multicolumn{3}{c}{(e) OT-GAN} \\

\includegraphics[width=1.3in]{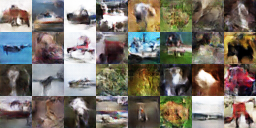} & \includegraphics[width=1.3in]{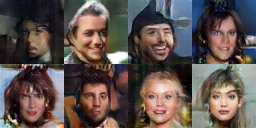} & \includegraphics[width=1.3in]{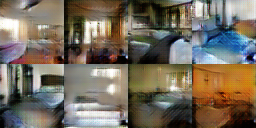} \\ \hline

\multicolumn{3}{c}{(f) Fr\'{e}chet-GAN} \\

\includegraphics[width=1.3in]{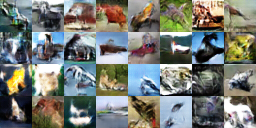} & \includegraphics[width=1.3in]{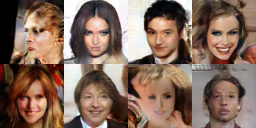} & \includegraphics[width=1.3in]{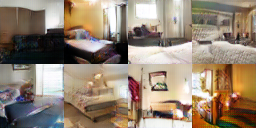} \\ \hline
 
\end{tabular}
\caption{Generated images of multi-channel datasets.}
\label{fig:generated_images_all}
\end{table*}

Furthermore, without BN, Fr\'{e}chet-GAN can generate digits that has a lesser degree of mode collapse, compared to the digits generated by SWG, and that are sharper, compared to Max-SWG and OT-GAN. In Table~\ref{tbl:fids_nobn}, we show the quantitative results of more compared approaches with and without BN in term of their FIDs. It is known that we can detect intra-class mode collapse through FID~\cite{borji2019pros}. For both MNIST and CIFAR-10, we observe that the values of FID increase significantly, especially in CIFAR-10, without BN. However, both FID's of Fr\'{e}chet-GAN and Max-SWG only slightly increase. This shows the robustness of Fr\'{e}chet-GAN with respect to the model architecture; note also, the significantly smaller values of FID achieved by Fr\'{e}chet-GAN (without BN). 

\renewcommand\tablename{Table}
\begin{table*}[!t]
\centering
\caption{FIDs of the methods.}
\begin{tabular}{l>{\raggedleft\arraybackslash}p{2.3cm}>{\raggedleft\arraybackslash}p{2.3cm}>{\raggedleft\arraybackslash}p{2.3cm}>{\raggedleft\arraybackslash}p{2.8cm}}
\hline
\hline
& \textbf{MNIST} & \textbf{CIFAR-10} & \textbf{CELEB-A} & \textbf{LSUN-Bedroom} \\

\hline
\hline

WGAN            & 22.44 $\pm$ 1.07      & 44.51 $\pm$ 0.98       & 24.49 $\pm$ 0.98         & 51.88 $\pm$ 3.21  \\
WGAN-GP         & 21.40 $\pm$ 3.29      & 36.24 $\pm$ 0.98       & 22.17 $\pm$ 1.09         & 48.98 $\pm$ 5.25  \\
SWG             & 16.89 $\pm$ 0.95      & 26.68 $\pm$ 0.80       & 10.93 $\pm$ 0.59         & 26.97 $\pm$ 3.42 \\
Max-SWG         & 15.24 $\pm$ 1.94      & \underline{23.56  $\pm$ 0.54}     & 10.13 $\pm$ 0.61      & 40.14 $\pm$ 4.51 \\
OT-GAN          & 18.63 $\pm$ 0.74      & 32.50 $\pm$ 0.64      & 19.40 $\pm$ 2.98          & 70.49 $\pm$ 5.25 \\
Fr\'{e}chet-GAN     & \underline{10.51 $\pm$ 2.73}       & 24.64 $\pm$ 0.54      & \underline{9.80 $\pm$ 0.78}   & \underline{16.34 $\pm$ 2.98} \\
\hline

\end{tabular}
\label{tbl:generated_images_fids}
\end{table*}

\subsection{Image Generation}

In this section, we provide the results of generating images of our proposed methods and the representative baselines from both the dual and primal domains. 

The generated images from different methods are presented in Figure~\ref{fig:generated_images_all}. The visual quality of the images generated by OT-GAN and Fr\'{e}chet-GAN is comparable to the existing GANs. Especially, Fr\'{e}chet-GAN can generate images with the minimal defective artifacts. For higher-dimensional images, as shown in Figure~\ref{fig:generated_images_high_resolution}, the images generated by Fr\'{e}chet-GAN are comparable to those of Max-SWG. In Table~\ref{tbl:generated_images_fids}, we present the quantitative metric, FID, of the methods. All primal-domain GANs have significant improvement over the dual-domain GANs. While OT-GAN's FIDs are worse than those of SWG and Max-SWG, Fr\'{e}chet-GAN achieves the best FIDs in MNIST, CELEB-A and LSUN-Bedroom datasets and a comparable FID to Max-SWG in CIFAR-10. This supports our claim in this paper: 
directly minimizing the Wasserstein distance in the feature space works for the image generative modeling task. Furthermore, Fr\'{e}chet distance is heuristically shown as a more robust cost function as a proxy for estimating the Wasserstein distance for use in such image generation applications. 

\begin{table*}[!t]
\renewcommand\tablename{Figure}
\centering
\begin{tabular}{lc}
\rotatebox{90}{Max-SWG} &
\includegraphics[width=4.5in]{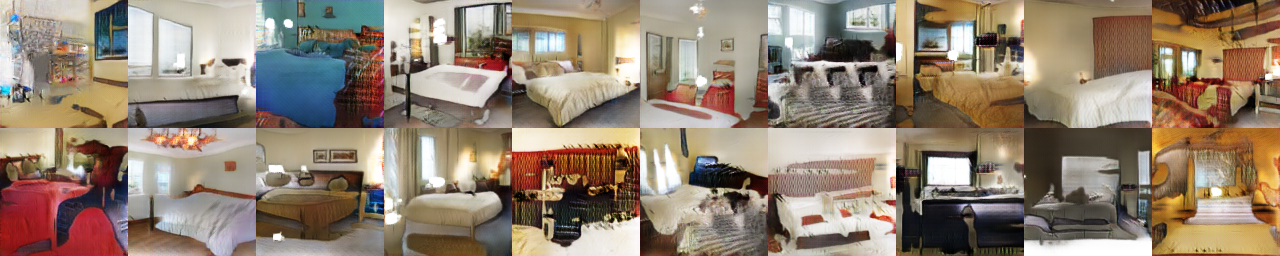} \\
\rotatebox{90}{Fr\'{e}chet-GAN} &
\includegraphics[width=4.5in]{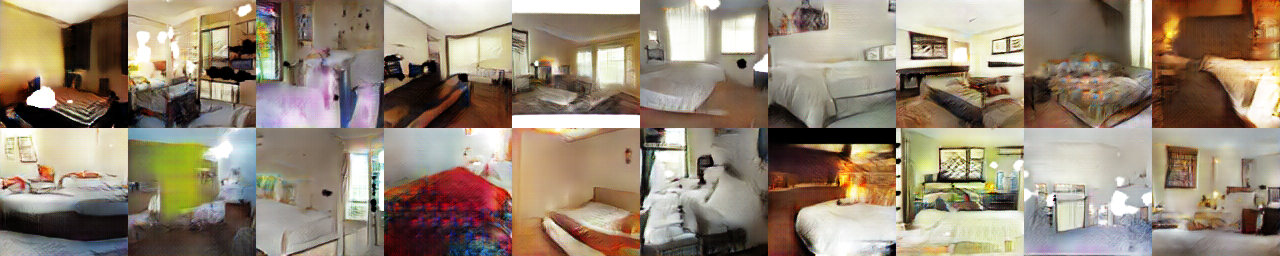} \\
\end{tabular}
\caption{High-resolution LSUN-bedroom Generated images (128x128).}
\label{fig:generated_images_high_resolution}
\vspace{-20pt}
\end{table*}

\section{Conclusion} \label{sec:conclusion}

In this paper, we propose new and efficient GAN approaches which estimate the distributional distance in the low-dimensional, feature space. The feature space is chosen from the GAN discriminator. We employ two different choices of the distributional distance, Optimal Transport and Fr\'{e}chet distance, which lead to the two new GANs, OT-GAN and Fr\'{e}chet-GAN, respectively. Our motivation of Fr\'{e}chet-GAN is from the fact that Fr\'{e}chet Inception Distance is a solid metric for evaluating the quality of the generated images and it is equivalent to the Wasserstein-2 formulation of the OT solution (but with a significant improvement in computational complexity) when the feature space distributions are multivariate Gaussians. Furthermore, we propose an efficient, differentible algorithm to calculate the Fr\'{e}chet distance. While it is previously argued that Wasserstein distance has an exponential sample complexity, it is heuristically shown that our proposed approaches to estimate the Wasserstein distance in the low-dimensional feature space  overcomes such complexity of the high-dimensional space. In fact, Fr\'{e}chet-GAN achieves a significant improvement in both the visual inspection of the generated images and FID, compared to the existing primal and dual GANs. The results motivate us to study other types of distributional distances in the feature space of GANs, especially in the image generation domain.



%
%
\bibliographystyle{splncs04}
\bibliography{main}

\begin{thebibliography}{10}
\providecommand{\url}[1]{\texttt{#1}}
\providecommand{\urlprefix}{URL }
\providecommand{\doi}[1]{https://doi.org/#1}

\bibitem{arjovsky2017towards}
Arjovsky, M., Bottou, L.: Towards principled methods for training generative
  adversarial networks. arxiv e-prints, art. arXiv preprint arXiv:1701.04862
  (2017)

\bibitem{bartels1972}
Bartels, R.H., Stewart, G.W.: Solution of the matrix equation ax + xb = c [f4].
  Commun. ACM  \textbf{15}(9),  820–826 (Sep 1972).
  \doi{10.1145/361573.361582}, \url{https://doi.org/10.1145/361573.361582}

\bibitem{blei2003latent}
Blei, D.M., Ng, A.Y., Jordan, M.I.: Latent dirichlet allocation. Journal of
  machine Learning research  \textbf{3}(Jan),  993--1022 (2003)

\bibitem{borji2019pros}
Borji, A.: Pros and cons of gan evaluation measures. Computer Vision and Image
  Understanding  \textbf{179},  41--65 (2019)

\bibitem{burkard2009assignment}
Burkard, R.E., Dell'Amico, M., Martello, S.: Assignment problems. Springer
  (2009)

\bibitem{denman1976matrix}
Denman, E.D., Beavers~Jr, A.N.: The matrix sign function and computations in
  systems. Applied mathematics and Computation  \textbf{2}(1),  63--94 (1976)

\bibitem{deshpande2019max}
Deshpande, I., Hu, Y.T., Sun, R., Pyrros, A., Siddiqui, N., Koyejo, S., Zhao,
  Z., Forsyth, D., Schwing, A.G.: Max-sliced wasserstein distance and its use
  for gans. In: Proceedings of the IEEE Conference on Computer Vision and
  Pattern Recognition. pp. 10648--10656 (2019)

\bibitem{deshpande2018generative}
Deshpande, I., Zhang, Z., Schwing, A.G.: Generative modeling using the sliced
  wasserstein distance. In: Proceedings of the IEEE conference on computer
  vision and pattern recognition. pp. 3483--3491 (2018)

\bibitem{doan2020hashing}
Doan, K., Kimiyaie, A., Manchanda, S., Reddy, C.K.: Image hashing by minimizing
  independent relaxed wasserstein distance. arXiv preprint arXiv:2003.00134
  (2020)

\bibitem{genevay2017learning}
Genevay, A., Peyr{\'e}, G., Cuturi, M.: Learning generative models with
  sinkhorn divergences. arXiv preprint arXiv:1706.00292  (2017)

\bibitem{goodfellow2014generative}
Goodfellow, I., Pouget-Abadie, J., Mirza, M., Xu, B., Warde-Farley, D., Ozair,
  S., Courville, A., Bengio, Y.: Generative adversarial nets. In: Advances in
  neural information processing systems. pp. 2672--2680 (2014)

\bibitem{gulrajani2017improved}
Gulrajani, I., Ahmed, F., Arjovsky, M., Dumoulin, V., Courville, A.C.: Improved
  training of wasserstein gans. In: Advances in neural information processing
  systems. pp. 5767--5777 (2017)

\bibitem{heusel2017gans}
Heusel, M., Ramsauer, H., Unterthiner, T., Nessler, B., Hochreiter, S.: Gans
  trained by a two time-scale update rule converge to a local nash equilibrium.
  In: Advances in neural information processing systems. pp. 6626--6637 (2017)

\bibitem{higham1997stable}
Higham, N.J.: Stable iterations for the matrix square root. Numerical
  Algorithms  \textbf{15}(2),  227--242 (1997)

\bibitem{iohara2018generative}
Iohara, A., Ogawa, T., Tanaka, T.: Generative model based on minimizing exact
  empirical wasserstein distance (2018),
  \url{https://openreview.net/forum?id=BJgTZ3C5FX}

\bibitem{ionescu2015matrix}
Ionescu, C., Vantzos, O., Sminchisescu, C.: Matrix backpropagation for deep
  networks with structured layers. In: Proceedings of the IEEE International
  Conference on Computer Vision. pp. 2965--2973 (2015)

\bibitem{krizhevsky2009learning}
Krizhevsky, A., Hinton, G., et~al.: Learning multiple layers of features from
  tiny images  (2009)

\bibitem{lecun1998mnist}
LeCun, Y.: The mnist database of handwritten digits. nec research institute
  (1998)

\bibitem{liu2015deep}
Liu, Z., Luo, P., Wang, X., Tang, X.: Deep learning face attributes in the
  wild. In: Proceedings of the IEEE international conference on computer
  vision. pp. 3730--3738 (2015)

\bibitem{manchanda2020regression}
Manchanda, S., Doan, D.K., Yadav, P., Keerthi, S.S.: Regression via implicit
  models and optimal transport cost minimization. arXiv preprint
  arXiv:2003.01296  (2020)

\bibitem{martin2017wasserstein}
Martin~Arjovsky, S., Bottou, L.: Wasserstein generative adversarial networks.
  In: Proceedings of the 34 th International Conference on Machine Learning,
  Sydney, Australia (2017)

\bibitem{metz2016unrolled}
Metz, L., Poole, B., Pfau, D., Sohl-Dickstein, J.: Unrolled generative
  adversarial networks. arXiv preprint arXiv:1611.02163  (2016)

\bibitem{miyato2018spectral}
Miyato, T., Kataoka, T., Koyama, M., Yoshida, Y.: Spectral normalization for
  generative adversarial networks. arXiv preprint arXiv:1802.05957  (2018)

\bibitem{salakhutdinov2007restricted}
Salakhutdinov, R., Mnih, A., Hinton, G.: Restricted boltzmann machines for
  collaborative filtering. In: Proceedings of the 24th international conference
  on Machine learning. pp. 791--798 (2007)

\bibitem{yu2015lsun}
Yu, F., Seff, A., Zhang, Y., Song, S., Funkhouser, T., Xiao, J.: Lsun:
  Construction of a large-scale image dataset using deep learning with humans
  in the loop. arXiv preprint arXiv:1506.03365  (2015)

\end{thebibliography}

\end{document}